\documentclass[sigconf,10pt,nonacm]{acmart}
\makeatletter
\@ACM@balancefalse
\renewcommand{\@ACM@checkaffil}{}
\makeatother

\AtBeginDocument{%
  }

\setcopyright{none}
\renewcommand\footnotetextcopyrightpermission[1]{}

\usepackage{pifont}
\usepackage{multirow}
\usepackage{booktabs}            
\usepackage{amsmath,amsfonts}
\usepackage{graphicx}
\usepackage{xcolor}
\usepackage{makecell}
\usepackage{enumitem}
\usepackage{xspace}
\usepackage{algorithm}           
\usepackage{algpseudocode}       
\usepackage[most]{tcolorbox}     
\usepackage{tikz}
\usepackage{subcaption}
\usepackage{circledsteps}
\usetikzlibrary{shapes.geometric,arrows.meta,positioning,fit,backgrounds,calc}

\newcommand{\smallstep}[1]{%
  {\scriptsize\Circled[inner xsep=5pt, inner ysep=5pt]{#1}}%
}

\providecommand{\up}{$\uparrow$}
\providecommand{\dn}{$\downarrow$}
\providecommand{\na}{--}
\providecommand{\placeholderfigure}[2]{%
\fbox{\parbox[c][#1][c]{0.94\linewidth}{\centering\small #2}}%
}
\providecommand{\legendplotvspace}{-6.5mm}

\newcommand{\system}{PastForward}

\title{\system{}: Faster On-Device GUI Agents via~Computational Experience Reuse}

\author{Taehwan Park}
\authornote{These authors contributed equally to this work.}
\affiliation{\institution{UNIST}}
\email{pth2002@unist.ac.kr}

\author{Changmin Lee}
\authornotemark[1]
\affiliation{\institution{UNIST}}
\email{lchm1106@unist.ac.kr}

\author{Hayeon Lee}
\affiliation{\institution{Meta Superintelligence Labs}}
\email{hayeonlee@meta.com}

\author{Taesik Gong}
\affiliation{\institution{UNIST}}
\email{taesik.gong@unist.ac.kr}

\renewcommand{\shortauthors}{Park et al.}

\begin{document}
\begin{abstract}
Running GUI agents on edge devices can keep sensitive screens and interaction histories local, but the computational cost of inference at every action step makes deployment challenging.
Existing GUI agent systems either perform full vision-language model (VLM) inference at each action step or reuse coarse-grained knowledge matched to prior tasks.
However, dynamic mobile environments and user tasks make it difficult to fully utilize prior task executions without additional fine-tuning or task-specific offline exploration.
To address this challenge, we present \textbf{\system{}}, a system that accelerates GUI agents through validated, fine-grained reuse of computational experience accumulated during ordinary task execution. During decoding, \system{} retrieves prior output sequences as device-adaptive multi-token proposals and verifies them in a single VLM forward pass. Across action steps, it uses prior GUI transitions to begin next-step inference while the device executes the current action, retains the early computation only when the predicted screen matches the observed screen, and carries reusable KV states forward.
We evaluate \system{} on AndroidWorld workloads derived from real mobile usage patterns using multiple VLM backbones across server and edge platforms. On device, \system{} achieves action-step latency speedups of $1.63\text{--}2.36\times$ while maintaining task success rates.
\end{abstract}


\maketitle
\begingroup
\renewcommand{\thefootnote}{}
\footnotetext{All experiments, data and model access, and processing activities were conducted by UNIST.}
\endgroup

\section{Introduction}
\label{sec:introduction}
Mobile GUI agents promise general-purpose smartphone automation: a user specifies an intent in natural language, and the system carries it out within or across applications~\cite{appagent,autodroid,mobilegpt,autodroidv2,verisafe,vdroid,agentprog,mobileagentv35,maiui,uitars,qin2026executable,awm}. Vision-language models (VLMs) extend mobile automation beyond application APIs and task-specific scripts by enabling agents to interpret application screens, reason about multi-step goals, and interact with graphical interfaces designed for human users~\cite{appagent,mobileagentv35,maiui,uitars,autorpa,mi2026darwinian}. Executing an agent task, however, requires a repeated interaction loop. At each action step, the agent processes the current screen, generates and executes an action, and waits for the application to expose the resulting screen. 


Many existing GUI agent systems rely on cloud-hosted models for these calls, requiring privacy-sensitive screen representations and interaction context to be sent to a remote service during execution~\cite{appagent,autodroid,mobilegpt,verisafe,autorpa,vdroid}.
Running the VLM locally on mobile or edge devices can keep these inputs local and remove remote communication from the loop, but shifts the inference workload to resource-constrained mobile devices~\cite{lu2025bluelm,vlmcache}. In our characterization on a Galaxy~S26 Ultra smartphone, VLM inference takes 8.57\,s per action step and accounts for 77.26\% of action-step latency, measured over the complete screen-to-screen cycle including both VLM inference and application processing. Because each action step depends on the preceding action and resulting screen, this computational cost of inference accumulates along the critical path of a multi-step task.

In practice, mobile use is concentrated in a small set of applications, and in-app actions recur over time~\cite{de2019strategies,liu2019characterizing}. Recent traces of real-world mobile task execution show a similarly concentrated application distribution~\cite{fingertip}. 
We find corresponding recurrence across GUI-agent task executions. Under the user-calibrated workload, previously observed VLM output sequences of at least four consecutive tokens cover 64.1\% of generated tokens, while GUI transitions recur in 15.1\% of action steps. Even under the task-once workload, where each task is instantiated only once, these rates remain 59.6\% and 11.2\%, respectively (\S\ref{sec:analysis}).
These measurements show that recurrence occurs at the level of token sequences and GUI transitions, even when no complete task or trajectory repeats.

Prior systems derive relatively coarse-grained app-specific knowledge, reusable subtasks, workflows, or action sequences from application exploration or past trajectories~\cite{autodroid,mobilegpt,mi2026darwinian,qin2026executable,autorpa,appagent,autodroidv2,awm}. Some of these systems can avoid VLM calls when a stored procedure is directly executable, but they do not reuse fine-grained computation within individual VLM calls.
At the model and runtime levels, existing techniques reduce the computational cost of agent inference through visual-token pruning or visual-computation reuse~\cite{fastv,vlmcache}. These techniques exploit structure within individual model invocations or across nearby screens, but do not treat execution records accumulated across task executions as a reusable resource for later inference.

However, such fine-grained execution records cannot be reused directly. A change in the instruction, task parameter, or screen may cause the current output to diverge from a previously generated token sequence. Likewise, even the same action from the same screen may produce a different next screen, for example, when a pop-up appears in one execution but not another. A prior output therefore cannot simply be replayed as the current action, and its intermediate computation cannot be treated as an exact cache hit. The central challenge is to identify which prior computation remains valid at each action step and reuse only that computation without altering the agent's decision.

We present \system{}, a system that addresses this challenge by treating execution records accumulated during task execution as computational experience and using them to accelerate VLM computation only to the extent validated by the current VLM or observed screen. During decoding, \emph{Experience-Guided Adaptive Decoding} retrieves prior token sequences as multi-token proposals, accepts only the matching prefix verified by the current VLM, and adapts the proposal length to observed acceptance and device-specific computational costs of inference. While the application is processing the current action, \emph{Early Inference via Next-Screen Prediction} uses a prior GUI transition to anticipate the next screen and begin VLM inference for the next action step before the observed screen becomes available. The early computation is retained only when the predicted and observed screens agree. \emph{Cross-Step State Carryover} additionally reuses KV states across action steps to reduce repeated prefill computation.
Together, these mechanisms reduce action-step latency through fine-grained reuse of computational experience, and the benefits grow as computational experience accumulates.
Moreover, \system{} requires no offline exploration or additional model fine-tuning, preserves the VLM backbone and agent policy, and operates on device.

We evaluate \system{} on the AndroidWorld~\cite{rawles2025androidworld} benchmark with 20 applications and 116 tasks using two workload distributions: task-uniform and user-calibrated.
The evaluation covers three VLM agent backbones and four hardware platforms: a server and three edge devices (Jetson Orin Nano, Galaxy S26 Ultra, and Xiaomi 15).
\system{} achieves $1.40\text{--}1.68\times$ and $1.63\text{--}2.36\times$ action-step speedups on the server and edge devices, respectively, while maintaining task success rates. 
\system{} also incurs little computational overhead in memory use: on the Jetson Orin Nano, it increases peak RAM usage by only 0.1\% (2.4\,MiB).



\section{Characterization and Motivation}
\label{sec:analysis}

\subsection{GUI Agent Action-Step Latency}
\label{sec:characterization_cost}

A conventional GUI agent alternates between VLM inference and application execution. Starting from the current screen $S_t$ at step $t$, the agent invokes a vision-language model (VLM) with the task context and $S_t$. Each VLM invocation performs prefill followed by autoregressive decoding and produces a VLM output sequence $\mathbf{y}_t$, from which the agent obtains action $a_t$. 
After sending $a_t$, the agent waits for the interface to settle before observing the resulting screen $S_{t+1}$~\cite{rawles2025androidworld,xie2024osworld}. We denote this observed GUI transition by $S_t \xrightarrow{a_t} S_{t+1}$ and refer to the complete cycle as an \emph{action step}. Each action step consists of VLM inference that produces $a_t$ and the action-to-observation (A2O) interval from $a_t$ to $S_{t+1}$. Under conventional execution, the next VLM invocation cannot begin until this A2O interval completes. Accordingly, \emph{action-step latency} measures the end-to-end latency of a single GUI agent cycle, from the arrival of the current screen $S_t$ to that of the resulting screen $S_{t+1}$. 

\begin{figure}[t]
    \centering

    \begin{subfigure}[t]{0.495\columnwidth}
        \vspace{0pt}
        \centering
        \includegraphics[width=\linewidth]{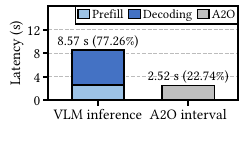}
        \caption{Latency breakdown.}
    \end{subfigure}\hspace{0.01\columnwidth}%
    \begin{subfigure}[t]{0.495\columnwidth}
        \vspace{0pt}
        \centering
        \includegraphics[width=\linewidth]{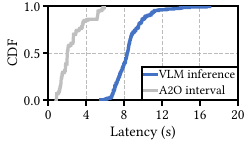}
        \caption{Latency distribution.}
    \end{subfigure}
    \vspace{-0.8mm}
    \caption{On-device measurements of GUI agent latency using Qwen3-VL-2B on a Galaxy~S26 Ultra.
    }
    \label{fig:physical_latency}
\end{figure}

The inference component is particularly computationally expensive on-device, where VLMs operate under substantially tighter resource constraints than in cloud deployments. Figure~\ref{fig:physical_latency} reports VLM inference and A2O latency on a Galaxy~S26 Ultra across 17 tasks spanning four applications.
Figure~\ref{fig:physical_latency}(a) decomposes the mean action step into 2.62\,s of prefill, 5.95\,s of decoding, and a 2.52\,s A2O interval. Together, prefill and decoding account for 77.26\% of action-step latency. Figure~\ref{fig:physical_latency}(b) shows the distribution behind those means: VLM inference latency has a median of 8.34\,s and a 90th percentile of 10.45\,s, while the A2O interval has a median of 2.01\,s and a 90th percentile of 5.20\,s. During conventional sequential execution, the VLM remains idle throughout the A2O interval while waiting for the resulting screen. Because multi-step tasks invoke the VLM repeatedly, inference and waiting time both accumulate along the task's critical path.

\subsection{Reuse Opportunities Across Task Executions}
\label{sec:characterization_reuse}

We characterize two forms of recurrence across task executions: recurring token sequences within VLM outputs and recurring GUI transitions $S_t \xrightarrow{a_t} S_{t+1}$.
For this characterization, we use AndroidWorld~\cite{rawles2025androidworld}, an interactive mobile agent benchmark that runs on a live Android emulator. It comprises 20 real-world Android applications and 116 programmatic tasks. Each task defines a parameterized workflow, and instantiating it with concrete values produces a task instance. We use \emph{task} to refer to one of the 116 programmatic tasks, \emph{task instance} to a particular parameterization, and \emph{task execution} to one run of the agent on a task instance. During a task execution, the agent interacts online with the running applications, so the resulting screens and transitions come from live application execution rather than a static trace.

\begin{figure}[t]
    \centering
    \begin{subfigure}[t]{0.49\columnwidth}
        \vspace{0pt}
        \centering
        \includegraphics[width=\linewidth]{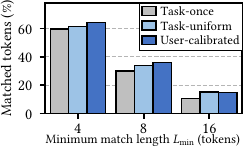}
        \caption{Token sequences.}
    \end{subfigure}
    \hfill
    \begin{subfigure}[t]{0.49\columnwidth}
        \vspace{0pt}
        \centering
        \includegraphics[width=\linewidth]{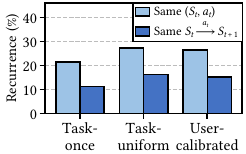}
        \caption{GUI transitions.}
    \end{subfigure}

    \vspace{-1mm}
    \caption{Recurrence across AndroidWorld workloads. (a) Fraction of
generated tokens covered by prior exact sequences at each minimum match
length $L_{\min}$. (b) Fraction of action steps matching a prior
$(S_t,a_t)$ pair and the subset that also reproduces $S_{t+1}$.
}
    \label{fig:recurrence}
\end{figure}


We construct three AndroidWorld workloads that separate recurrence across different tasks from recurrence caused by repeatedly selecting the same task. The \emph{task-once} workload executes one instance of each of the 116 tasks. For each random seed, we construct two repeated-use workloads, each consisting of 348 task instances. The \emph{task-uniform} workload samples tasks uniformly, whereas the \emph{user-calibrated} workload matches the application-level concentration observed in real-world user demonstrations of mobile agent tasks~\cite{fingertip}. Whenever a task is selected again, we instantiate it with new parameter values, so no exact task instance is replayed. \S\ref{sec:eval:setup} details the workload construction and calibration.


\noindent\textbf{Token-sequence recurrence.}
We first quantify recurrence within VLM output sequences. For each VLM output sequence, we measure the fraction of generated tokens covered by non-overlapping exact token sequences observed in prior output sequences, excluding tokens fixed by the structured tool-call format. Figure~\ref{fig:recurrence}(a) shows that token-sequence recurrence arises even when each task is instantiated only once. In the task-once workload, previously observed sequences cover 59.6\% of generated tokens at $L_{\min}=4$. Repeated-use workloads provide slightly higher coverage, reaching 61.4\% under task-uniform and 64.1\% under user-calibrated at $L_{\min}=4$. Exact sequence matches persist across all three workloads even at longer sequence lengths.
Recurring token sequences therefore provide an opportunity to reuse computation from prior outputs and thus reduce VLM inference latency.

\noindent\textbf{Transition recurrence.}
For each action step $S_t \xrightarrow{a_t} S_{t+1}$, we first ask whether its $(S_t,a_t)$ pair matches that of a transition observed in an earlier task execution, and then whether the resulting $S_{t+1}$ also matches. As shown in Figure~\ref{fig:recurrence}(b), prior $(S_t,a_t)$ matches occur in 21.4\%, 27.2\%, and 26.4\% of action steps under the task-once, task-uniform, and user-calibrated workloads, respectively. Conditioned on a prior $(S_t,a_t)$ match, the same $S_{t+1}$ reappears in 52.3\%, 59.2\%, and 57.2\% of cases. Across all action steps, GUI transitions recur in 11.2\%, 16.1\%, and 15.1\% of cases. Thus, a recurring $(S_t,a_t)$ context is informative about, but does not determine, the resulting $S_{t+1}$.
Recurring GUI transitions similarly provide an opportunity to reuse prior interaction outcomes and reduce action-step latency.

\subsection{Design Challenges and Implications}
\label{sec:characterization_implications}
The measurements above expose three design challenges that must be addressed to turn recurrence into latency reduction.

\noindent\textbf{Recurrence capture below the task level.}
The recurring units in \S\ref{sec:characterization_reuse} are token sequences and individual GUI transitions, rather than complete tasks or trajectories. Either can recur in another task execution, so reuse should be indexed by local decoding or interaction context without requiring a task- or trajectory-level match. This fine-grained approach can maximize reuse opportunities.

\noindent\textbf{Robustness to divergence from prior task executions.}
Neither form of recurrence establishes that prior results remain valid for the current task execution. A retrieved token sequence may agree with the current VLM output only up to a point of divergence, while a previously observed transition may lead to a different screen under the current application state. A reuse mechanism must therefore benefit from recurring computation without assuming that prior and current task executions remain identical.

\noindent\textbf{Runtime computational overhead and device-aware adaptation.}
Recurrence creates opportunities for reuse, but exploiting them adds computation and memory use. Whether reuse reduces latency depends on the available opportunities and the compute and memory constraints of the target device. A practical design should therefore keep its own computational overhead low and account for device-specific execution conditions.

\section{\system{} Design}
\subsection{Overview}
\label{sec:design_overview}

\begin{figure}[t]
    \centering
    \includegraphics[width=1\columnwidth]
        {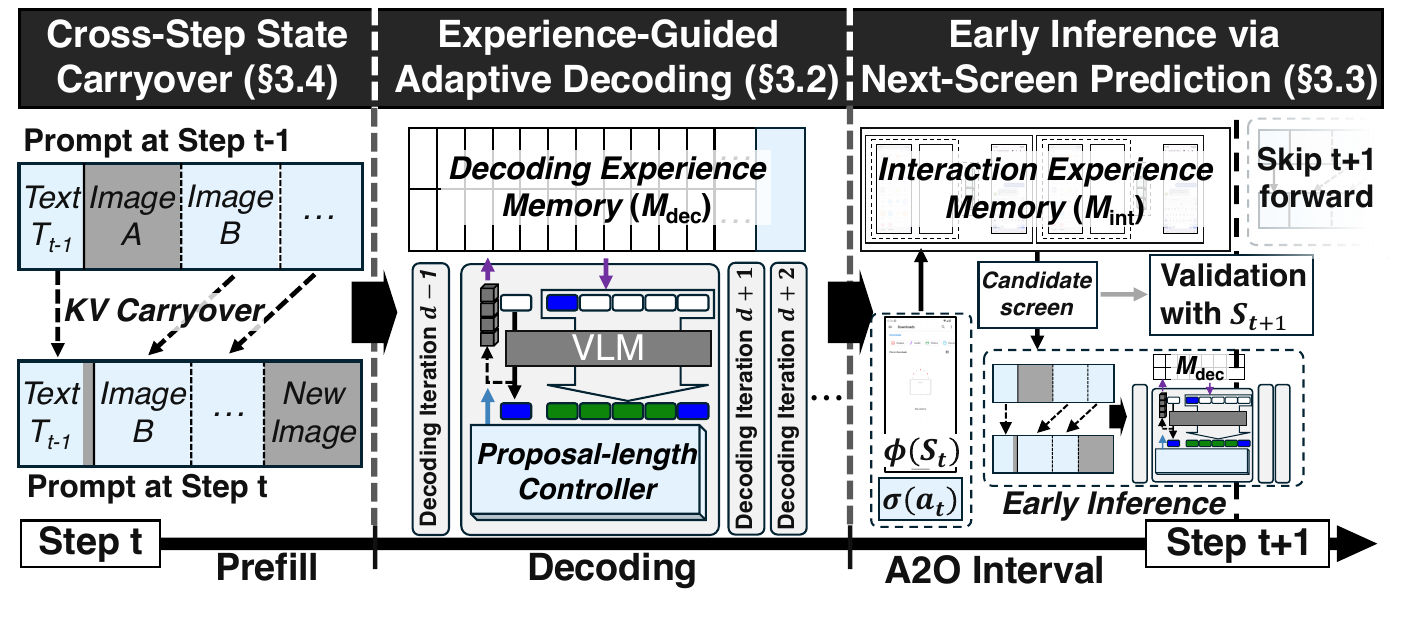}
    \caption{Overview of \system{}.
    }
    \label{fig:overview}
\end{figure}

Figure~\ref{fig:overview} provides an overview of \system{}. Building on the characterization in \S\ref{sec:analysis}, \system{} uses computational experience accumulated across task executions to address two major sources of GUI agent latency: VLM inference and the A2O interval during application execution. It reuses this experience only when it is validated against the current VLM's predictions or the observed screen.

\emph{Experience-Guided Adaptive Decoding} (\S\ref{sec:tokensequence_reuse}) reuses prior token sequences as multi-token proposals, allowing a single VLM forward pass to verify and accept multiple tokens. It dynamically adapts the proposal length to the acceptance behavior and computational cost of verification observed at runtime. \emph{Early Inference via Next-Screen Prediction} (\S\ref{sec:shadow_execution}) uses prior GUI transitions to predict the next screen and starts VLM inference for the next action step on the predicted screen during the current action's A2O interval, reusing the early computation only if the observed screen matches the prediction. Furthermore, \emph{Cross-Step State Carryover} (\S\ref{sec:runtime_state}) reduces repeated prefill work by carrying shared KV state across consecutive action steps.

\subsection{Experience-Guided Adaptive Decoding}

\begin{figure}
    \centering
    \includegraphics[width=1\linewidth]{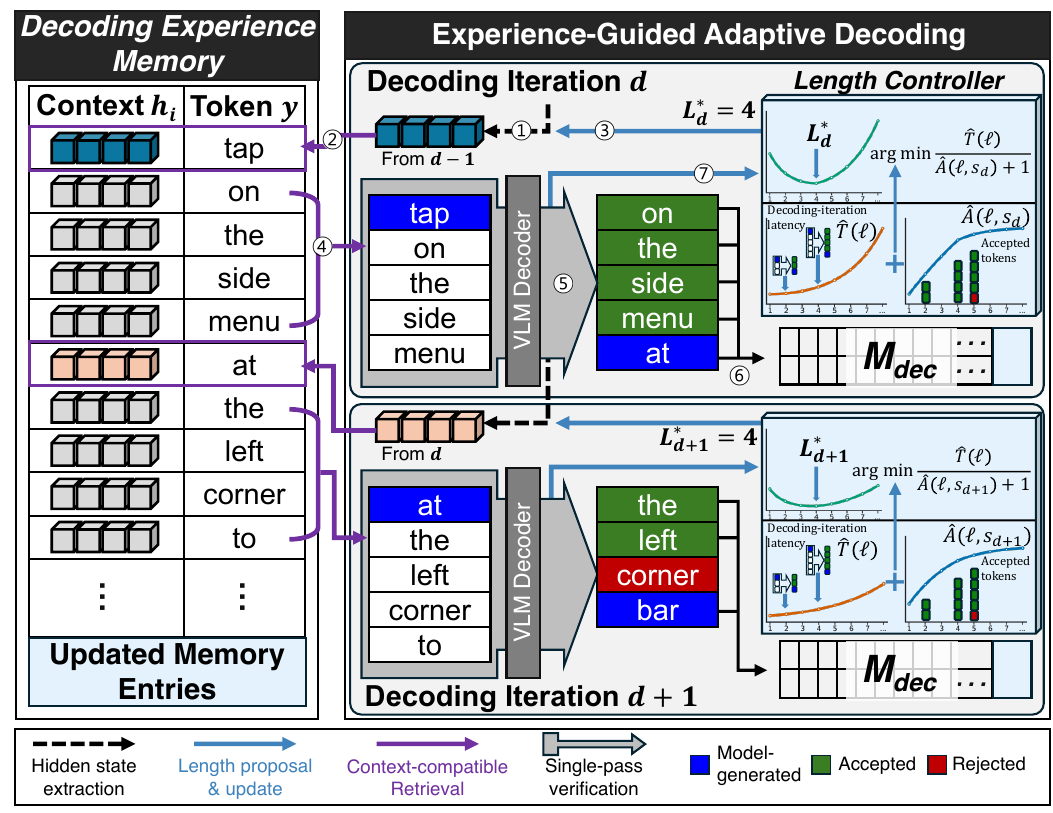}
    \caption{Workflow of Experience-Guided Adaptive Decoding. 
    }
    \label{fig:adaptive_decoding}
\end{figure}

As investigated in \S\ref{sec:characterization_reuse}, GUI agents frequently revisit similar screen contexts and generate token sequences that recur across action steps and task executions. Conventional autoregressive decoding does not exploit this computational recurrence and advances an output sequence by one token per sequential forward pass. \system{} instead treats previously generated output sequences as reusable decoding experience: it retrieves a context-compatible memory entry, adaptively chooses how many tokens to propose, and verifies the resulting proposal with the VLM in one forward pass. We call each retrieval-and-verification cycle a \emph{decoding iteration}, indexed by $d$. The process repeats until the output sequence is complete, with each updated context used to retrieve the next proposal. Figure~\ref{fig:adaptive_decoding} shows two consecutive iterations.

\label{sec:tokensequence_reuse}

\noindent\textbf{Decoding experience memory.}
To reuse these token sequences, \system{} stores prior VLM output sequences in the Decoding Experience Memory, $\mathcal{M}_{\mathrm{dec}}$, and indexes them at individual token positions.
At each action step, the VLM generates an output token sequence $\mathbf{y}=(y_1,\ldots,y_T)$ through autoregressive decoding, where $T$ is the sequence length. At token position $i$, the context $h_i$ denotes the decoder's final-layer state from which the VLM predicts $y_i$. The same token can appear under different multimodal prompts and generated output prefixes, so token identity alone is an ambiguous retrieval key. $h_i$, however, is conditioned on both the multimodal prompt and $\mathbf{y}_{<i}$.
\system{} therefore indexes decoding experience by $h_i$. Conceptually, $\mathcal{M}_{\mathrm{dec}}$ associates $h_i$ with the suffix $\mathbf{y}_{i:T}$ at every position $i<T$. During retrieval, the stored $y_i$ must match the current token; the remaining suffix $\mathbf{y}_{i+1:T}$ then serves as a candidate token sequence. Storing every suffix separately would duplicate tokens. Instead, \system{} appends each output sequence once to a shared token array and stores each entry as $(h_i,p_i,b_i)$, where $p_i$ is the position of $y_i$ and $b_i$ is its \emph{output-sequence boundary}, the last position of the same output sequence. A proposal of length $\ell$ can therefore access only positions $p_i+1,\ldots,\min(p_i+\ell,b_i)$, preventing retrieval from crossing into the next action-step output. For instance, in $[A,B,C \mid D,E]$, an entry pointing to token $B$ has its output-sequence boundary at token $C$, so its proposal may include $C$ but cannot extend to $D$. $\mathcal{M}_{\mathrm{dec}}$ persists across task executions and supports cosine-similarity lookup.

\noindent\textbf{Context-compatible retrieval.}
At each decoding iteration, \system{} searches $\mathcal{M}_{\mathrm{dec}}$ using the current context and the token predicted from that state. For the first iteration of an action step, the final-layer state at the final prompt position produces the current VLM's first greedy prediction. Together, this state and token form the initial \emph{retrieval query}; subsequent queries use the pair $(h_i,y_i)$ produced by the preceding greedy or verification pass (\smallstep{1}). 
Given a query $(h_i,y_i)$, \system{} first restricts the search to stored entries $(h_j,p_j,b_j)$ whose token at position $p_j$ matches $y_i$ and for which $p_j<b_j$. It then maps $h_i$ into the key space of $\mathcal{M}_{\mathrm{dec}}$ and selects the eligible entry with the highest cosine similarity, obtaining its similarity score $s_t$ (\smallstep{2}). The proposal length $\ell$ is then selected at runtime, as described later in this section (\smallstep{3}). The selected entry supplies up to $\ell$ subsequent tokens as a proposal, truncated at the stored output sequence's boundary (\smallstep{4}). If no entry is eligible, the iteration uses an ordinary greedy pass.

\noindent\textbf{Single-pass token verification.}
When retrieval returns a proposal, \system{} verifies its tokens against the current VLM before accepting them. Given the current token $y_i$ and proposal $\hat{\mathbf{y}}_{i+1:i+\ell}$, a single VLM forward pass computes predictions at every proposed position and at one additional position following the proposal (\smallstep{5}). Let $\ell_{\mathrm{acc}}$ be the length of the longest proposal prefix that matches these predictions. The iteration accepts this prefix and appends the model prediction at the first mismatch, or the prediction following the proposal if all $\ell$ tokens match, advancing the output by $\ell_{\mathrm{acc}}+1$ tokens. The final token appended during the iteration is produced directly by the current VLM; this token and the context that produced it form the next retrieval query. Because every accepted proposal token matches the current VLM's greedy prediction, \system{} produces the same output sequence as greedy decoding and incurs \emph{no accuracy loss}. This verification process resembles the parallel check used in speculative decoding~\cite{speculative_decoding}; however, rather than relying on a separate draft model, \system{} retrieves proposals from token sequences generated during prior agent executions and stored in $\mathcal{M}_{\mathrm{dec}}$. After each decoding iteration, \system{} adds the tokens appended during the iteration and their corresponding contexts to $\mathcal{M}_{\mathrm{dec}}$ (\smallstep{6}).

\noindent\textbf{Runtime-adaptive proposal-length controller.}
The efficiency of single-pass verification depends on the proposal length. A longer proposal creates more opportunities to accept multiple tokens in one VLM forward pass, but it also increases verification latency and wastes computation on positions after the first mismatch. A shorter proposal reduces this computational cost of verification and limits such wasted computation, but it caps the number of tokens that can be accepted per pass and may therefore require more decoding iterations. Furthermore, this balance shifts across devices and operating conditions because the computational cost of verifying each proposal length depends on both hardware capabilities and the device's runtime state.
\system{} therefore adapts the length at each decoding iteration using two online estimators. The \emph{acceptance estimator} predicts how many proposal tokens will be accepted at each candidate length and retrieval similarity, while the \emph{latency estimator} predicts the corresponding decoding-iteration latency on the current device. 
Given an eligible retrieval with similarity $s_d$, the controller selects the length with the lowest estimated latency (\smallstep{3}):

\begin{equation}
    L_d^{*}
    =
    \operatorname*{arg\,min}_{0\leq \ell \leq \ell_{\max}}
    \frac{\widehat{T}(\ell)}
         {\widehat{A}(\ell,s_d)+1},
    \label{eq:adaptive_length_objective} 
\end{equation}

Here, $\widehat{T}(\ell)$ is the estimated decoding-iteration latency, and $\widehat{A}(\ell,s_d)$ is the expected number of accepted proposal tokens with the similarity score $s_d$. The additional $1$ accounts for the greedy prediction appended by every decoding iteration.
$\ell_{\max}$ is the configured upper bound on the proposal length; retrieved proposals are further truncated at their output-sequence boundaries. When the proposal length is selected as zero, the current decoding iteration performs an ordinary greedy pass.

After each decoding iteration, the controller updates both estimators using the observations (\smallstep{7}). The acceptance estimator uses online logistic regressions to estimate the probability of accepting the first proposal token and the conditional continuation probability for subsequent tokens, based on proposal length and retrieval similarity. The latency estimator fits an online log-quadratic function to the measured decoding-iteration latency as a function of proposal length. The controller normally selects the proposal length that minimizes the objective and occasionally explores an adjacent length. The controller adapts to retrieval quality and device latency by updating both estimators online.


\subsection{Early Inference via Next-Screen Prediction}
\label{sec:shadow_execution}
\begin{figure}
    \centering
    \includegraphics[width=1\linewidth]{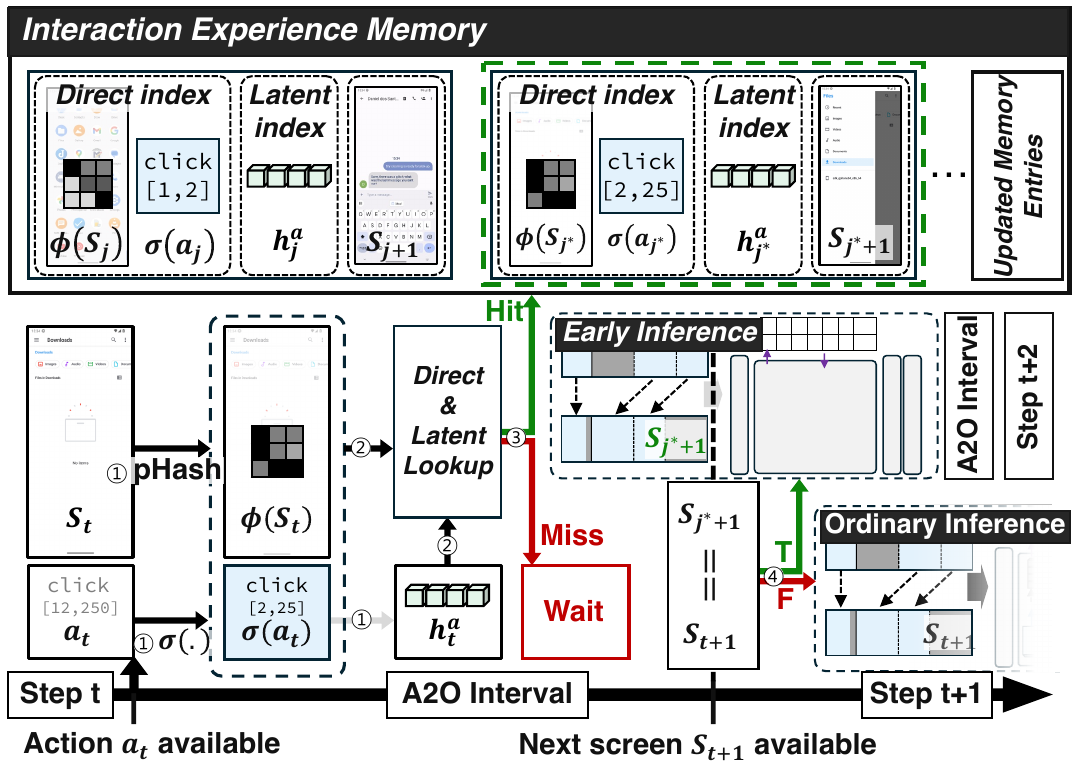}
    \caption{Workflow of Early Inference via Next-Screen Prediction.
    }
    \label{fig:early_inference}
\end{figure}

As explained in \S\ref{sec:characterization_cost}, conventional GUI agents leave the VLM idle during the action-to-observation (A2O) interval. After producing $a_t$, the agent must wait for the application to expose the resulting screen $S_{t+1}$ before invoking the VLM for the next action step. The recurring GUI transitions characterized in \S\ref{sec:characterization_reuse} allow \system{} to use this interval for next-step VLM inference. It retrieves the resulting screen of a prior transition as a candidate next screen and begins inference while the application processes $a_t$. Since a recurring interaction context does not guarantee the same next screen, \system{} retains the early computation only if the observed $S_{t+1}$ matches the candidate. Figure~\ref{fig:early_inference} shows when early inference begins and how its result is validated.

\noindent\textbf{Interaction experience memory.}
To retrieve a candidate next screen from prior task executions, \system{} stores each completed GUI transition $S_j \xrightarrow{a_j} S_{j+1}$ as a record $(\phi(S_j),\allowbreak \sigma(a_j),\allowbreak h_j^{a},\allowbreak S_{j+1})$ in the Interaction Experience Memory, $\mathcal{M}_{\mathrm{int}}$. $\mathcal{M}_{\mathrm{int}}$ maintains two indices over these records: a \emph{direct index} and a \emph{latent index}. 

The \emph{direct index} uses $\bigl(\phi(S_j),\sigma(a_j)\bigr)$ as its key, where $\phi$ maps a screen to a perceptual hash (pHash)~\cite{zauner2010implementation} and $\sigma$ encodes the action type and the arguments that determine its effect. A pHash is a compact visual fingerprint designed so that visually similar screens produce similar values. The direct index returns a record only when both the screen hash and action signature match those of a stored record. This prevents reusing a transition for the same action on a different screen or for a different action on the same screen. However, a direct-key match is sufficient but not necessary: non-identical but similar screens (e.g., because of changes in app content) can still lead to the same next screen when the corresponding actions are generated under similar contexts. 

To capture such cases, the \emph{latent index} indexes each record by $h_j^{a}$, enabling fallback retrieval when the direct index misses, subject to a minimum cosine-similarity threshold. Here, the action-generation representation $h_j^{a}$ summarizes the model context in which $a_j$ was generated. Let $\mathcal{I}_j^{a}$ denote the token positions spanning the generated tool call. \system{} computes $h_j^{a}=\frac{1}{|\mathcal{I}_j^{a}|}\sum_{i\in\mathcal{I}_j^{a}} h_i$. Unlike $\mathcal{M}_{\mathrm{dec}}$, which indexes individual token positions, this mean-pooled representation summarizes the complete tool call.

Both indices reference the same record and therefore its resulting screen $S_{j+1}$. Their keys are computed when $a_j$ is generated, but the record is completed only after the application exposes $S_{j+1}$. Thus, $\mathcal{M}_{\mathrm{int}}$ contains only transitions observed during actual execution. Concrete choices for the perceptual hash, action-coordinate normalization, latent-key projection, and similarity threshold are described in \S\ref{sec:implementation}.

\noindent\textbf{Next-screen retrieval.}
Given the current screen $S_t$ and action $a_t$, \system{} constructs the direct lookup key $(\phi(S_t),\sigma(a_t))$, while $h_t^a$ serves as the latent lookup query (\smallstep{1}). \system{} first queries the direct index with this key; if the exact lookup misses, it queries the latent index with $h_t^a$ (\smallstep{2}). The latent path can retrieve a transition generated under a similar action context even when its exact screen-action key differs. Let $j^{*}$ denote the index of the record returned by either lookup. A hit returns its stored next screen $S_{j^{*}+1}$ as the candidate; if both indices miss, \system{} waits for the observed screen and skips early inference for the current action step (\smallstep{3}).

\noindent\textbf{Early inference and validation during A2O interval.}
Once $S_{j^{*}+1}$ is retrieved, \system{} constructs the input for the next action step from the current task instruction and interaction history. Because the actual next screen $S_{t+1}$ has not yet been observed, \system{} temporarily uses the retrieved candidate $S_{j^{*}+1}$ in its place. Thus, $S_{j^{*}+1}$ is the only substituted component of the next-step input and the only component checked during validation.
Using the candidate screen, \system{} begins the standard next-step VLM inference. Retrieval and inference therefore proceed during the A2O interval, hiding the portion completed before $S_{t+1}$ becomes available. Any unfinished computation remains on the critical path, while any completed output is withheld until the observed screen validates the candidate.

When the resulting screen $S_{t+1}$ becomes available, \system{} compares its application content with that of the retrieved candidate $S_{j^{*}+1}$. All computation and runtime state produced by early inference remain isolated until this validation completes. If the screens match, \system{} uses the early inference computation for the next action step; otherwise, it discards the early-inference state and runs ordinary VLM inference using the observed screen $S_{t+1}$ (\smallstep{4}). Because early computation is used only after an exact application-content match, the validated next-step input is the same as in ordinary inference, so the early-inference mechanism incurs no accuracy loss. No action produced by early inference is sent to the application before the candidate screen has been validated.

\subsection{Cross-Step State Carryover}
\label{sec:runtime_state}

Conventional prefix caching reuses KV states only along the longest identical prompt prefix~\cite{zheng2024sglang}. GUI agent prompts, however, evolve at every action step. Moreover, recent GUI agents retain multiple previous screenshots so that each decision can use recent interaction history~\cite{mobileagentv35,mobileagentv3,maiui,ui-hawk,showui}. As the agent appends the latest action and screenshot, a rolling multi-image window may evict the oldest screenshot and shift every retained image to a new token position. The retained visual content is unchanged, but its tokens now fall outside the reusable prefix. \system{} therefore handles the unchanged prefix, retained screenshots, and newly introduced content separately to extend KV-state reuse and reduce prefill latency.

\noindent\textbf{Prompt-prefix KV-state carryover.}
Prompt-prefix KV-state carryover reuses the KV states of tokens in the unchanged prefix shared by consecutive action-step prompts to reduce repeated computation during prefill.
Let $P_t$ and $P_{t+1}$ denote the tokenized multimodal prompts at consecutive action steps within the same task execution, and let $p$ be the length of their longest identical prefix. Tokens within this prefix retain the same content, positions, and causal context. \system{} therefore recycles their KV states and resumes computation from the first position after the shared prefix.

\noindent\textbf{Rolling-image KV-state carryover.}
Rolling-image KV-state carryover reuses the KV states of screenshots retained across consecutive action steps to reduce repeated visual-token computation during prefill.
Let $W$ denote the number of screenshots in the window. When the window advances from $[I_r,\ldots,I_{r+W-1}]$ to $[I_{r+1},\ldots,I_{r+W}]$, the retained images move to new positions in the multimodal prompt. 
Without additional handling, conventional prefix caching cannot reuse the KV states of these shifted visual tokens, so the VLM must recompute them during prefill. To avoid this recomputation, \system{} maps each retained visual-token block to its new span and adjusts its cached keys, $\widetilde{K}_{m}$, from position $m$ to $m'$ using $\widetilde{K}_{m'}=R(m')R(m)^{-1}\widetilde{K}_{m}$, where $R(m)$ is the model's rotary position embedding (RoPE) rotation at token position $m$~\cite{roformer,qwen2vl}. The corresponding value states are reused unchanged. Position adjustment corrects the positional encoding but cannot reconstruct a changed causal context. A retained screenshot may have originally been encoded with an evicted screenshot and earlier action history to its left, so rolling-image KV-state carryover remains approximate. \system{} applies it only to retained screenshots; the newly observed screen, changed text, and remaining prompt suffix are computed normally.

\section{Implementation}
\label{sec:implementation}

\noindent\textbf{Runtime integration.}
We implement \system{} as a common inference runtime supporting multiple VLM-based GUI agents in AndroidWorld~\cite{rawles2025androidworld}.
The integration preserves each agent's prompt construction, image processing, VLM output parsing, and action interface.

\noindent\textbf{Memory and retrieval.} \system{} maps normalized final-layer hidden states to 256-dimensional int8 keys through a fixed random projection, and caps proposal length at \mbox{$\ell_{\max}=64$}. The Interaction Experience Memory uses a hash table keyed by a 64-bit perceptual screen hash and an action signature, quantizes coordinates into fixed-size grid cells, and applies a cosine-similarity threshold of 0.98 to latent retrieval. Screens are stored locally, while interaction records retain their file paths and content hashes.

\noindent\textbf{Runtime state management.} \system{} runs Early Inference in a background worker whose KV cache, decoding memory, and proposal-controller state remain isolated until screen validation completes. The runtime uses a contiguous BF16 cache for both prompt-prefix KV-state carryover and rolling-image KV-state carryover, as described in \S\ref{sec:runtime_state}.


\providecommand{\tbd}{\textbf{TBD}}
\providecommand{\up}{$\uparrow$}
\providecommand{\dn}{$\downarrow$}
\providecommand{\na}{--}
\providecommand{\placeholderfigure}[2]{%
\fbox{\parbox[c][#1][c]{0.94\linewidth}{\centering\small #2}}%
}
\providecommand{\legendplotvspace}{-5.2mm}
\providecommand{\panellegendplotvspace}{-1.5mm}

\section{Evaluation}
\label{sec:eval}

\subsection{Evaluation Setup}
\label{sec:eval:setup}

\noindent\textbf{Benchmark.}
We evaluate on AndroidWorld~\cite{rawles2025androidworld}, an interactive live-emulator benchmark comprising 116 parameterized, multi-step tasks across 20 real-world Android applications, and use the task terminology defined in \S\ref{sec:characterization_reuse}.

\begin{figure}[t]
    \centering
    \includegraphics[width=\linewidth]{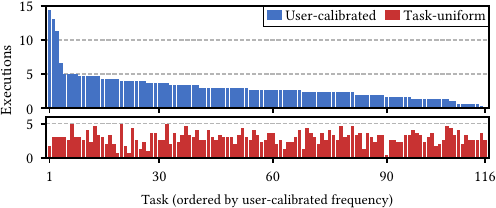}
    \caption{
    Number of executions for each task under the two workloads ($M=348$ task instances).
    }
    \label{fig:workload}
\end{figure}

\noindent\textbf{Workload construction.}
We use FingerTip20K~\cite{fingertip}, a mobile GUI agent benchmark built from 21,437 task demonstrations collected longitudinally from 95 users on their own phones, to calibrate our user-calibrated workload.
Unlike passive app-usage traces, FingerTip20K records users' intents as they arise during daily use and the corresponding action sequences. Its app distribution therefore provides a proxy for how user-initiated mobile tasks are concentrated across applications.

The $N=116$ tasks in AndroidWorld are distributed unequally across the $K=20$ applications. Let $n_i$ denote the number of tasks associated with application $i$. 
We sample an app-level distribution
\(\boldsymbol{\pi} \sim \operatorname{Dirichlet}\!\left(
\alpha K \left[\frac{n_1}{N},\ldots,\frac{n_K}{N}\right]
\right)\) for each random seed
and construct a sequence of $M$ \emph{task instances} by sampling an application according to $\boldsymbol{\pi}$ and uniformly selecting one of its tasks. Repeatedly selected tasks are instantiated with new parameters (e.g., phone number) each time, so no exact task instance is repeated within a workload. Because $\mathbb{E}[\pi_i]=n_i/N$, $\alpha$ controls app-level concentration without changing the expected application mix. As $\alpha\rightarrow\infty$, this construction becomes uniform over tasks rather than applications. We approximate this limit with $\alpha=10^6$ and refer to it as the \emph{task-uniform} workload. We calibrate $\alpha$ for the \emph{user-calibrated} workload against the app-level concentration observed in FingerTip20K. 
In an offline sweep of $\alpha$, $\alpha=1.0$ most closely matches this concentration, so we use it for the user-calibrated workload.  

Figure~\ref{fig:workload} shows the realized task distributions under the user-calibrated and task-uniform workloads, averaged over three random seeds. We use $M=348$ task instances, equivalent to three instances per benchmark task, for the experiments in \S\ref{sec:eval:e2e}--\S\ref{sec:eval:experience}.
For the design alternatives and system computational overhead analyses in \S\ref{sec:eval:Design_alternative}--\S\ref{sec:eval:overhead}, we use the user-calibrated workload with $M=116$ task instances.

\begin{table}[t]
\centering
\caption{Evaluation platforms. SD8 denotes Snapdragon 8; memory denotes server GPU memory or edge-device system memory.}
\label{tab:hardware_platforms}
\footnotesize
\setlength{\tabcolsep}{2pt}
\renewcommand{\arraystretch}{1.10}
\resizebox{\columnwidth}{!}{%
\begin{tabular}{@{}llll@{}}
\toprule
Platform & CPU/SoC & GPU/NPU & Memory \\
\midrule
Server & 2$\times$ EPYC 9554 & RTX PRO 6000 / \na & 96\,GB GDDR7 \\
Jetson Orin Nano & 6-core Cortex-A78AE & Ampere / \na & 8\,GB LPDDR5 \\
Galaxy S26 Ultra & SD8 Elite Gen 5 & Adreno/Hexagon & 16\,GB LPDDR5X \\
Xiaomi 15 & SD8 Elite & Adreno/Hexagon & 12\,GB LPDDR5X \\
\bottomrule
\end{tabular}%
}
\end{table}

\noindent\textbf{Platforms and VLM deployment.} Table~\ref{tab:hardware_platforms} summarizes our evaluation platforms and their hardware specifications. For server-side experiments, we run Qwen3-VL-8B-Instruct~\cite{bai2025qwen3}, GUI-Owl-1.5-8B-Instruct~\cite{mobileagentv35}, and MAI-UI-8B~\cite{maiui}, with the AndroidWorld environment, Android emulator, and VLM inference on the same host. 
For on-device experiments, we run Qwen3-VL-2B-Instruct~\cite{bai2025qwen3} on a Jetson Orin Nano, a Samsung Galaxy S26 Ultra, and a Xiaomi 15, while the AndroidWorld environment and emulator run on a separate controller host. This setup preserves AndroidWorld's emulator-based task execution while evaluating VLM inference on target devices. On the Jetson, we load BF16 weights directly in the agent process using Hugging Face Transformers and PyTorch/CUDA. On the smartphones, we use Q8\_0-quantized GGUF language-model weights and an FP16 multimodal projector with a custom Android llama.cpp runtime, which maps prefill and decoding to the NPU. Reported on-device action-step latency includes network communication.

\noindent\textbf{Baselines.} We compare \system{} with Vanilla, MobileGPT~\cite{mobilegpt}, V-Droid~\cite{vdroid}, FastV~\cite{fastv}, and Agent-X~\cite{agentx}. Vanilla follows the official inference configuration of each agent backbone. MobileGPT uses offline exploration to prepopulate its app memory with reusable subtasks for later recall. V-Droid constructs a discrete action space and uses a separately trained 8B verifier to select the next action rather than generating it autoregressively. Both MobileGPT and V-Droid rely on cloud-hosted LLMs. For reproducible local evaluation, we replace their proprietary cloud LLM calls with Qwen3.5-122B-A10B (FP8)~\cite{qwen3.5} served by vLLM; the MobileGPT baseline uses its original exploration and memory pipeline, while the V-Droid baseline uses its original 4-bit Llama-3.1-8B verifier. 
FastV is a representative token pruning method that retains the top 50\% of visual tokens after the first two decoder layers. Agent-X is a recent method that accelerates LLM-based agents through prompt restructuring, prefix caching, and speculative decoding. For Agent-X, we evaluate only ExSpec using a trigram lookup table and a draft length of four. To adapt its per-query lookup table to our multi-step GUI execution setup, we retain the table across action steps and task instances and update it with model outputs. We exclude Agent-X's PromptWeaver component because it restructures TinyAgent's textual prompt~\cite{tinyagent}, whereas we preserve each backbone's native multimodal prompt. 
All methods use standard KV caching for each inference call.

\noindent\textbf{Metrics.} We report task success rate (SR), prefill latency, decoding latency, and action-step latency. SR is the fraction of task instances completed successfully. Prefill and decoding latencies are averaged over all action steps. Action-step latency is measured from the arrival of the current screen to that of the resulting screen after action execution and thus includes all latencies: prefill, experience retrieval, decoding, and the A2O interval. 

\subsection{Overall System Performance}
\label{sec:eval:e2e}

\newcommand{\pmv}[1]{{\scriptsize$\pm$#1}}

\begin{table*}[t]
\centering
\caption{
Server-side performance on AndroidWorld under the task-uniform and user-calibrated workloads.
SR and latency values are reported as mean$\pm$s.d. across three random seeds.
Values in parentheses report action-step-latency speedup over Vanilla.
\textsuperscript{\dag} For MobileGPT, prefill includes the entire LLM call due to its non-streaming API.
}
\label{tab:e2e_server}

\small
\setlength{\tabcolsep}{2.8pt}

\resizebox{\textwidth}{!}{%
\begin{tabular}{llcccccccc}
\toprule

\multirow{2}{*}{Model(s)}
& \multirow{2}{*}{Method}
& \multicolumn{4}{c}{Task-uniform}
& \multicolumn{4}{c}{User-calibrated}
\\
\cmidrule(lr){3-6}\cmidrule(lr){7-10}

& & SR (\%) & Prefill (s) & Decoding (s) & Action step (s)
  & SR (\%) & Prefill (s) & Decoding (s) & Action step (s)
\\
\midrule
Qwen3.5-122B-A10B
& MobileGPT
& 26.34\pmv{1.78} & 2.851\pmv{0.416} & --\textsuperscript{\dag} & 5.15\pmv{0.41}
& 23.90\pmv{2.28} & 2.700\pmv{0.515} & --\textsuperscript{\dag} & 5.02\pmv{0.79} \\
\addlinespace[5pt]
\begin{tabular}[c]{@{}l@{}}
Qwen3.5-122B-A10B \\
+ Llama-3.1-8B verifier
\end{tabular}
& V-Droid
& 43.58\pmv{1.76} & 1.692\pmv{0.190} & 0.321\pmv{0.007} & 4.02\pmv{0.16}
& 37.21\pmv{8.75} & 1.957\pmv{0.693} & 0.329\pmv{0.020} & 4.21\pmv{0.77} \\
\midrule
\multirow{4}{*}{Qwen3-VL-8B}
& Vanilla
& 46.12\pmv{1.01} & 0.447\pmv{0.004} & 2.443\pmv{0.017} & 5.62\pmv{0.20}
& 42.91\pmv{6.33} & 0.452\pmv{0.007} & 2.465\pmv{0.121} & 5.52\pmv{0.14}
\\
& FastV
& 41.62\pmv{1.30} & \textbf{0.420\pmv{0.004}} & 2.421\pmv{0.016} & 5.47\pmv{0.16} (1.03$\times$)
& 40.71\pmv{4.81} & \textbf{0.425\pmv{0.006}} & 2.463\pmv{0.090} & 5.49\pmv{0.12} (1.01$\times$)
\\
& Agent-X
& 45.98\pmv{2.24} & 0.454\pmv{0.003} & 1.252\pmv{0.002} & 4.44\pmv{0.08} (1.26$\times$)
& 44.78\pmv{4.97} & 0.458\pmv{0.008} & 1.253\pmv{0.025} & 4.30\pmv{0.14} (1.28$\times$)
\\
& \textbf{\system{}}
& 47.56\pmv{2.01} & 0.445\pmv{0.002} & \textbf{0.659\pmv{0.020}} & \textbf{3.66\pmv{0.18} (1.54$\times$)}
& 43.58\pmv{5.20} & 0.447\pmv{0.001} & \textbf{0.640\pmv{0.012}} & \textbf{3.39\pmv{0.28} (1.63$\times$)}
\\
\midrule
\multirow{4}{*}{MAI-UI-8B}
& Vanilla
& 56.56\pmv{1.17} & 1.206\pmv{0.013} & 2.111\pmv{0.045} & 5.98\pmv{0.12}
& 52.49\pmv{5.64} & 1.254\pmv{0.089} & 2.152\pmv{0.064} & 5.97\pmv{0.11}
\\
& FastV
& 53.64\pmv{1.36} & 1.151\pmv{0.041} & 2.000\pmv{0.046} & 5.79\pmv{0.07} (1.03$\times$)
& 52.39\pmv{3.02} & 1.217\pmv{0.103} & 1.998\pmv{0.144} & 5.76\pmv{0.12} (1.04$\times$)
\\
& Agent-X
& 54.79\pmv{0.44} & 1.204\pmv{0.011} & 1.177\pmv{0.043} & 5.05\pmv{0.06} (1.18$\times$)
& 52.59\pmv{5.23} & 1.261\pmv{0.079} & 1.186\pmv{0.056} & 5.00\pmv{0.07} (1.19$\times$)
\\
& \textbf{\system{}}
& 55.12\pmv{2.20} & \textbf{0.824\pmv{0.003}} & \textbf{0.623\pmv{0.022}} & \textbf{3.71\pmv{0.04} (1.61$\times$)}
& 52.73\pmv{6.32} & \textbf{0.839\pmv{0.032}} & \textbf{0.613\pmv{0.056}} & \textbf{3.56\pmv{0.11} (1.68$\times$)}
\\
\midrule
\multirow{4}{*}{GUI-Owl-1.5-8B}
& Vanilla
& 55.82\pmv{0.30} & 1.674\pmv{0.026} & 1.213\pmv{0.006} & 6.85\pmv{0.08}
& 46.70\pmv{2.24} & 1.764\pmv{0.086} & 1.235\pmv{0.023} & 6.71\pmv{0.13}
\\
& FastV
& 57.90\pmv{2.23} & 1.562\pmv{0.005} & 1.105\pmv{0.018} & 6.53\pmv{0.14} (1.05$\times$)
& 52.20\pmv{4.35} & 1.616\pmv{0.082} & 1.092\pmv{0.010} & 6.46\pmv{0.09} (1.04$\times$)
\\
& Agent-X
& 56.66\pmv{1.06} & 1.665\pmv{0.036} & 0.565\pmv{0.012} & 6.09\pmv{0.14} (1.13$\times$)
& 50.19\pmv{5.47} & 1.734\pmv{0.082} & 0.544\pmv{0.019} & 6.00\pmv{0.13} (1.12$\times$)
\\
& \textbf{\system{}}
& 57.81\pmv{1.36} & \textbf{1.027\pmv{0.008}} & \textbf{0.237\pmv{0.006}} & \textbf{4.88\pmv{0.28} (1.40$\times$)}
& 51.96\pmv{4.70} & \textbf{1.052\pmv{0.012}} & \textbf{0.225\pmv{0.012}} & \textbf{4.55\pmv{0.44} (1.47$\times$)}
\\
\bottomrule
\end{tabular}%
}
\end{table*}

\noindent\textbf{Server-side performance.} 
Table~\ref{tab:e2e_server} shows that \system{} achieves action-step speedups of $1.40\text{--}1.61\times$ over Vanilla under the task-uniform workload and $1.47\text{--}1.68\times$ under the user-calibrated workload.
Qwen3-VL benefits mainly from adaptive decoding because its prompt contains only the current screenshot, whereas MAI-UI and GUI-Owl retain three and five screenshots, respectively, enabling rolling-image KV-state carryover to reduce prefill latency.
\system{} achieves a decoding speedup of $3.39\text{--}5.49\times$ over Vanilla and $1.89\text{--}2.42\times$ over Agent-X. On MAI-UI and GUI-Owl, \system{}'s prefill speedup over FastV is $1.40\text{--}1.54\times$. 
FastV remains faster than \system{} in Qwen3-VL prefill, where rolling-image KV-state carryover is unavailable. 

\system{} and Vanilla achieve similar SR, with overlapping mean$\pm$s.d. ranges in most settings. The larger s.d. under the user-calibrated workload reflects task-mix variation, while AndroidWorld itself exhibits run-to-run variation even with the same seed~\cite{rawles2025androidworld}. Thus, the small SR differences need not indicate systematic accuracy loss, consistent with \system{}'s safeguards: single-pass token verification preserves greedy decoding outputs, and early inference is discarded on screen mismatches.
Across both workloads, the VLM-based GUI agents achieve higher SR than our reproduced configurations of MobileGPT and V-Droid. The two baselines use textual screen representations derived from Android accessibility information, whereas the VLM-based agents use screenshots directly. 

\begin{table}[t]
\centering
\caption{
Performance across three edge devices with Qwen3-VL-2B and the user-calibrated workload.
}
\label{tab:e2e_device}

\small
\setlength{\tabcolsep}{3.0pt}

\begin{tabular}{lcccc}
\toprule
Method & SR (\%) & Prefill (s) & Decoding (s) & Action step (s) \\
\midrule
\multicolumn{5}{l}{\emph{Jetson Orin Nano}} \\
Vanilla
& 21.84 & 1.964 & 14.388 & 19.71 \\
FastV
& 17.82 & 1.877 & 14.018 & 19.33 (1.02$\times$) \\
Agent-X
& 24.14 & 1.969 & 7.344 & 12.67 (1.56$\times$) \\
\textbf{\system{}}
& 21.70 & \textbf{1.597} & \textbf{4.227} & \textbf{8.36 (2.36$\times$)} \\
\midrule
\multicolumn{5}{l}{\emph{Galaxy S26 Ultra}} \\
Vanilla
& 19.40 & 2.837 & 5.131 & 10.86 \\
FastV
& 11.78 & 2.755 & 4.492 &  9.87 (1.10$\times$) \\
Agent-X
& 18.10 & 2.678 & 2.927 & 8.54 (1.27$\times$) \\
\textbf{\system{}}
& 18.10 & \textbf{2.282} & \textbf{2.037} & \textbf{6.66 (1.63$\times$)} \\
\midrule
\multicolumn{5}{l}{\emph{Xiaomi 15}} \\
Vanilla
& 17.39 & 4.066 & 8.160 & 15.11 \\
FastV
& 14.08 & 4.015 & 7.499 & 14.49 (1.04$\times$) \\
Agent-X
& 18.10 & 4.013 & 4.534 & 11.28 (1.34$\times$) \\
\textbf{\system{}}
& 17.53 & \textbf{3.179} & \textbf{3.091} & \textbf{9.17 (1.65$\times$)} \\
\bottomrule
\end{tabular}
\end{table}

\noindent\textbf{On-device performance.}
Table~\ref{tab:e2e_device} shows that \system{} achieves the lowest action-step latency across the Jetson Orin Nano, Galaxy S26 Ultra, and Xiaomi~15, with speedups of $1.63\text{--}2.36\times$ over Vanilla while maintaining SR. These end-to-end gains reflect reductions in both VLM inference phases: relative to Vanilla, prefill latency decreases by 18.7--21.8\% and decoding latency by 60.3--70.6\% across the three devices.
Relative to FastV and Agent-X, \system{} achieves action-step speedups of 1.48--2.31$\times$ and 1.23--1.52$\times$, respectively. These results show that \system{}'s benefits extend across Jetson CUDA and smartphone NPU runtimes.

\system{} provides similar relative decoding gains on the server and Jetson because both use BF16 PyTorch/CUDA kernels that process longer verification blocks efficiently. Decoding gains on smartphones are smaller because the computational costs of verification
rise more steeply with proposal length, leading the length controller to select shorter proposals. In contrast, \system{} provides larger relative prefill gains on the Jetson and smartphones than on the server. The server processes long prompt blocks efficiently with greater compute capacity and memory bandwidth, whereas prompt processing on these devices incurs comparatively computationally expensive transformer computation, KV-cache writes, and memory traffic, which can be avoided via \system{}'s Cross-Step State Carryover.

\subsection{Latency Reduction Breakdown}
\label{sec:eval:latency}
\begin{figure}[t]
    \centering
    \makebox[\linewidth][c]{%
      \includegraphics[width=\linewidth]{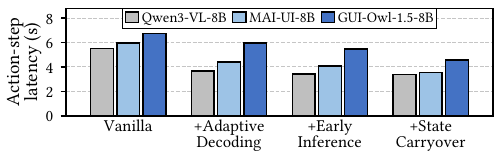}%
    }
    \caption{Action-step latency for three VLM backbones as \system{} components are added cumulatively to Vanilla under the user-calibrated workload.
    }
    \label{fig:cumulative}
\end{figure}

Figure~\ref{fig:cumulative} shows cumulative action-step latency reductions as \system{}'s three mechanisms (\S\ref{sec:tokensequence_reuse}--\S\ref{sec:runtime_state}) are incrementally added to Vanilla under the user-calibrated workload.

\noindent\textbf{Adaptive Decoding.} Experience-Guided Adaptive Decoding contributes the largest incremental reduction in action-step latency in Figure~\ref{fig:cumulative}, reducing decoding latency by $70.3\text{--}81.6\%$ across the three backbones. 
\system{} verifies each multi-token proposal against the current VLM's greedy predictions and accepts only the longest matching prefix, thereby preserving the greedy decoding output.

\noindent\textbf{Early inference.} Figure~\ref{fig:cumulative} shows the additional latency reduction from running next-step VLM inference during the A2O interval. To isolate this contribution, we count only the inference time hidden by matched next-screen predictions as latency savings and subtract the computational overhead of launching and resolving all Early Inference attempts. Averaged over all action steps, matched predictions hide $0.184\text{--}0.417$\,s of useful inference, yielding net action-step latency reductions of $0.154\text{--}0.389$\,s across the three backbones.

\begin{figure}[t]
    \centering
    \makebox[\linewidth][c]{%
      \includegraphics[width=\linewidth]{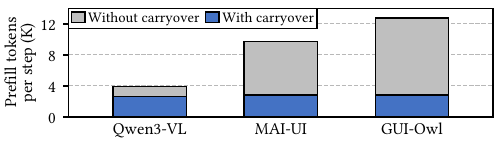}%
    }
    \caption{Number of tokens processed during prefill per action step with and without Cross-Step State Carryover.
    }
    \label{fig:carryover_tokens}
\end{figure}

\noindent\textbf{Cross-Step State Carryover.}
Figure~\ref{fig:carryover_tokens} shows how Cross-Step State Carryover reduces prefill work. For Qwen3-VL, the number of tokens processed during prefill decreases from 3.90K to 2.66K, but prefill latency changes only from 0.452\,s to 0.447\,s. Because its prompt contains only the current screenshot, carryover can reuse KV states for the unchanged prompt prefix but cannot reuse image KV states across action steps. MAI-UI and GUI-Owl retain multiple screenshots, enabling rolling-image KV-state carryover that reduces their prefilled token counts by $70.7\text{--}77.9\%$ and prefill latency by $31.7\text{--}40.4\%$.

\subsection{Effect of Experience Accumulation}
\label{sec:eval:experience}

\begin{figure}[t]
  \centering
  
  \makebox[\linewidth][c]{%
    \begin{minipage}[t]{0.50\linewidth}
      \centering
      \includegraphics[width=\linewidth]
        {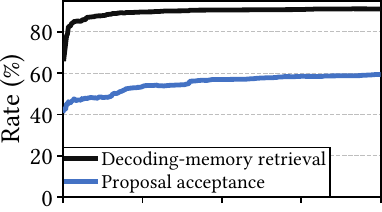}
    \end{minipage}%
    \hspace{1.0mm}
    \begin{minipage}[t]{0.50\linewidth}
      \centering
      \includegraphics[width=0.92\linewidth]
        {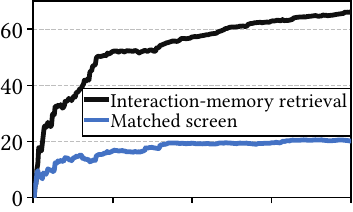}
    \end{minipage}%
  }

  \par\vspace{1mm}
  \makebox[\linewidth][c]{%
    \begin{subfigure}[t]{0.56\linewidth}
      \centering
      \hspace*{0.7mm}%
      \includegraphics[width=1\linewidth]
        {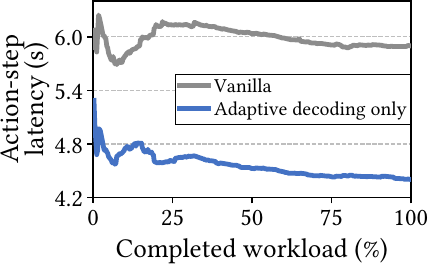}
      \hspace*{-0.7mm}%
      \caption{Adaptive decoding only.}
    \end{subfigure}%
    \begin{subfigure}[t]{0.56\linewidth}
      \centering
      \hspace*{2.1mm}%
      \includegraphics[width=\linewidth]
        {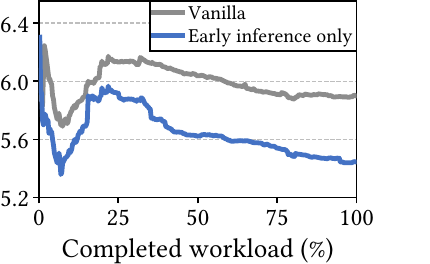}
      \hspace*{-2.1mm}%
      \caption{Early inference only.}
    \end{subfigure}%
  }
  \caption{Cumulative experience-reuse rates and action-step latency as computational experience accumulates under the user-calibrated workload with MAI-UI-8B.
  }
  \label{fig:experience_ablation}
\end{figure}

Figure~\ref{fig:experience_ablation} evaluates \system{}'s adaptive decoding and early inference as computational experience accumulates. To isolate their effects, we evaluate each on top of Vanilla with all other \system{} mechanisms disabled.

\noindent\textbf{Benefit of Adaptive Decoding over time.} Figure~\ref{fig:experience_ablation}(a) shows how the benefit of adaptive decoding changes as experience accumulates. We define the \emph{decoding-memory retrieval rate} as the fraction of decoding iterations in which a proposal is retrieved. The \emph{proposal acceptance rate} is the fraction of retrieved proposals that contribute at least one accepted token and thereby reduce latency. 
As decoding experience accumulates, the retrieval rate increases from 67.0\% to 91.1\%, while the proposal-acceptance rate follows a similar trend and reaches 59.4\% at completion.  
Thus, the accumulated memory entries provide more opportunities to reduce decoding latency through single-pass verification. The figure below reports the corresponding latency benefit. Relative to Vanilla, adaptive decoding reduces mean action-step latency by 1.037\,s over the first 10\% of task instances and by 1.606\,s over the last 10\%, yielding a 0.569\,s larger reduction late in the run. These net reductions show that the latency benefit of adaptive decoding grows as experience accumulates.


\noindent\textbf{Benefit of Early Inference over time.} Figure~\ref{fig:experience_ablation}(b) shows how the benefit of Early Inference changes as experience accumulates. We define the \emph{interaction-memory retrieval rate} as the fraction of action steps for which $\mathcal{M}_{\mathrm{int}}$ returns a next-screen candidate and the \emph{matched-screen rate} as the fraction of all action steps for which the retrieved candidate matches the observed next screen. As interaction experience accumulates, the interaction-memory retrieval rate increases to 66.1\% and the matched-screen rate converges to 20.2\% at completion. The figure below reports the corresponding latency benefit. Relative to Vanilla, Early Inference reduces mean action-step latency by 0.272\,s over the first 10\% of task instances. Over the last 10\%, the reduction grows to 0.577\,s. The larger late-run reduction is consistent with the mechanism: as the matched-screen rate grows, more action steps benefit from next-step VLM inference performed in advance during the A2O interval.

\subsection{Analysis of Design Alternatives}
\label{sec:eval:Design_alternative}

\begin{figure}[t]
      \centering
      \makebox[\linewidth][c]{%
        \begin{subfigure}[t]{0.50\linewidth}
          \centering
          \includegraphics[width=\linewidth]
            {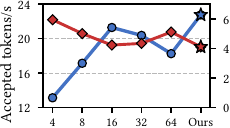}
          \subcaption{Jetson Orin Nano.}
          \label{fig:proposal_depth_jetson}
        \end{subfigure}%
        \par\vspace{1.0mm}
        \begin{subfigure}[t]{0.49\linewidth}
          \centering
          \includegraphics[width=\linewidth]
            {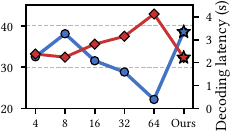}
          \subcaption{Galaxy S26 Ultra.}
          \label{fig:proposal_depth_galaxy}
        \end{subfigure}%
      }
      \makebox[\linewidth][l]{%
    \smash{%
        \raisebox{29.1mm}{%
            \makebox[\linewidth][c]{%
                \includegraphics[width=0.50\linewidth]
                    {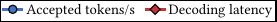}%
            }%
        }%
    }%
}

      \vspace{0.5mm}
      \caption{
      Comparison of fixed proposal lengths and our proposal-length controller on two edge devices. $\star$ marks the best value for each metric.
      }
      \label{fig:proposal_depth_devices}
\end{figure}

\noindent\textbf{Proposal-length adaptation.} 
Figure~\ref{fig:proposal_depth_devices} compares our runtime-adaptive proposal-length controller in \S\ref{sec:tokensequence_reuse} with fixed lengths $\ell\in\{4,8,16,32,64\}$ on the Jetson Orin Nano and Galaxy S26 Ultra, with retrieval and verification held constant. The fixed-length sweep reveals a device-dependent tradeoff. Decoding latency is minimized at $\ell=16$ on the Jetson and $\ell=8$ on the Galaxy. Beyond these lengths, the additional computational cost of verifying longer proposals outweighs the benefit of accepting more tokens, causing latency to increase. By tracking token acceptance and verification latency online, the controller selects the best proposal length for each device.

\begin{figure}[t]
      \centering
       \makebox[\linewidth][r]{%
        \makebox[0.5\linewidth][c]{%
        \includegraphics[width=0.5\linewidth]
            {figures/fig_operating_mode_sensitivity_legend.pdf}%
         }%
        }
      \makebox[\linewidth][c]{%
        \begin{subfigure}[t]{0.47\linewidth}
          \centering
          \includegraphics[width=\linewidth]
              {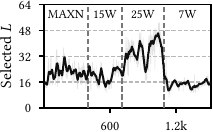}
          \subcaption{Online adaptation.}
          \label{fig:operating_mode_jetson_trace}
        \end{subfigure}%
        \par\hspace{1.0mm}
        \begin{subfigure}[t]{0.53\linewidth}
          \centering
          \raisebox{0.7mm}{
          \includegraphics[width=\linewidth]
              {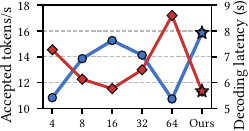}}
          \subcaption{Performance comparison.}
          \label{fig:operating_mode_jetson}
        \end{subfigure}%
      }

      \caption{
      Proposal-length adaptation under changing power modes on a
      Jetson Orin Nano.
      (a) Proposal length selected online by \system{}. Dashed
      lines mark mode transitions.
      (b) Corresponding results compared with
      fixed proposal lengths. $\star$ marks the best value. 
      }
\label{fig:operating_mode_sensitivity}
\end{figure}


Figure~\ref{fig:operating_mode_sensitivity} evaluates the adaptability of the proposal-length controller under changing power modes on a Jetson Orin Nano. Specifically, we use four \texttt{nvpmodel} profiles. The evaluated modes are MAXN\_SUPER (MAXN), 25\,W, 15\,W, and 7\,W.
Relative to MAXN, the 25\,W and 15\,W profiles retain six CPU cores and four GPU TPCs but lower their clock limits; the 7\,W profile further reduces the allocation to four CPU cores and two GPU TPCs. The EMC frequency limit is 3.199\,GHz for MAXN and 25\,W, and 2.133\,GHz for 15\,W and 7\,W.
The shorter proposal length under MAXN indicates that little reusable decoding experience has accumulated. Because the controller tracks token acceptance and verification latency online, subsequent changes reflect both retrieval quality and power-mode-specific latency. This adaptation yields better performance than fixed-length policies.


\begin{figure}[t]
    \centering
    \makebox[\linewidth][c]{%
      \includegraphics[width=\linewidth]{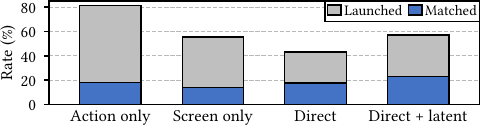}%
    }
    \caption{
Next-screen lookup-key ablation under the user-calibrated workload.
Gray bars show the launch rate; blue bars show the matched-screen rate.
}
    \label{fig:key_ablation}
\end{figure}

\noindent\textbf{Next-screen retrieval design.} Figure~\ref{fig:key_ablation} compares the launch and matched-screen rates for action-only, screen-only, direct, and \system{}'s two-stage direct-plus-latent lookups in~\S\ref{sec:shadow_execution}. Action-only and screen-only lookups launch Early Inference on 81.6\% and 55.3\% of action steps but match the observed next screen on only 18.2\% and 13.7\%, respectively, because each omits part of the transition context. Using both the screen and action yields a 43.1\% launch rate while increasing the matched-screen rate over screen-only lookup to 17.6\%.
Compared with direct lookup, \system{}'s hybrid design increases the matched-screen rate from 17.6\% to 23.0\%, while increasing the launch rate from 43.1\% to 57.3\%. The higher launch rate adds only the small computational cost of a mismatch quantified in~\S\ref{sec:eval:overhead}.

\subsection{System Computational Overhead}\label{sec:eval:overhead}

\begin{figure}[t]
      \centering
      \makebox[\linewidth][c]{%
        \begin{subfigure}[t]{0.5\linewidth}
          \centering
          \includegraphics[width=\linewidth]
            {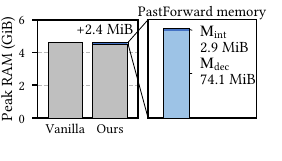}
          \subcaption{Peak RAM usage.}
        \end{subfigure}%
        \begin{subfigure}[t]{0.5\linewidth}
          \centering
          \includegraphics[width=\linewidth]
            {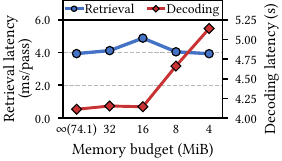}
          \subcaption{Memory-budget ablation.}
        \end{subfigure}%
      }
      \vspace{1mm}
      \caption{
      Computational overhead in memory use and decoding-memory budget tradeoff on the Jetson Orin Nano. (a) Peak RAM usage comparison with Vanilla. (b) Retrieval and decoding latency under different decoding memory budgets.
      }
      \label{fig:system_overhead_fig}
\end{figure}

\noindent\textbf{Runtime memory and storage.}
Figure~\ref{fig:system_overhead_fig}(a) compares the peak RAM usage of Vanilla and \system{}. \system{} increases peak RAM usage by 0.1\% (2.4\,MiB), from 4709.6\,MiB to 4712.0\,MiB. The Decoding Experience Memory, $\mathcal{M}_{\mathrm{dec}}$, and the Interaction Experience Memory, $\mathcal{M}_{\mathrm{int}}$, occupy 74.1\,MiB and 2.9\,MiB of RAM, respectively. Despite their combined 77.0\,MiB footprint, peak RAM increases by only 2.4\,MiB. This is because Cross-Step State Carryover reduces prefill activation memory by reusing KV states for prompt prefixes shared across steps.
Screens associated with interaction records are stored as PNG files on disk instead of RAM and occupy 291\,MiB at the end of the workload.
Because $\mathcal{M}_{\mathrm{dec}}$ accounts for most of the computational overhead in RAM use, Figure~\ref{fig:system_overhead_fig}(b) evaluates its latency tradeoff with an uncapped budget and fixed budgets.
Across all budgets, retrieval latency remains below 5\,ms per lookup and is relatively insensitive to total memory size. This stability comes from a lightweight token index that, for each query $(h_i,y_i)$, restricts cosine-similarity ranking to the compact decoding-state keys of entries whose stored token matches $y_i$ and whose position precedes its output-sequence boundary.
Reducing the budget from uncapped to 4\,MiB, however, increases per-step decoding latency from 4.111\,s to 5.137\,s because evicting entries from $\mathcal{M}_{\mathrm{dec}}$ reduces opportunities for token-sequence reuse. We leave the budget for $\mathcal{M}_{\mathrm{dec}}$ uncapped in the main evaluation.

\noindent\textbf{Online runtime computational overhead.}
\system{}'s online operations take 0.078, 0.089, and 0.043\,s per action step for Qwen3-VL-8B, MAI-UI-8B, and GUI-Owl-1.5-8B, respectively, accounting for at most 2.5\% of action-step latency. These operations include decoding-memory retrieval and updates, interaction-memory lookup and updates, and Early Inference launch and resolution. A next-screen mismatch adds $18.1\text{--}24.9$\,ms for screen validation, worker termination, and state rollback before inference restarts with the observed screen. The speedups in Table~\ref{tab:e2e_server} include these computational costs.

\section{Discussion}

\noindent\textbf{Compatibility with other VLM inference optimizations.} \system{} relies primarily on prior VLM inference reuse and next-step inference during the A2O interval. 
Other VLM inference optimizations, such as quantization, optimized attention kernels, visual-token pruning, and KV-cache compression, reduce the computational costs of both ordinary and early VLM inference and are largely complementary to \system{}.
However, if an optimization changes the hidden-state representation or KV layout, the affected model-state records must be rebuilt before they can be reused safely.

\noindent\textbf{Applicability and limitations.} \system{} reduces latency on the evaluated benchmark workloads. The observed gains appear to arise mainly from recurring decoding contexts and predictable screen transitions. However, the opportunities available in practice may vary by workload. 
For instance, infrequent recurrence among decoding contexts may limit the benefit of Experience-Guided Adaptive Decoding. Early Inference via Next-Screen Prediction could also have fewer opportunities to reduce latency when resulting screens vary substantially, or the A2O interval is too short to hide much computation.
In such cases, fewer proposals would be available for token verification, more early computation could be discarded during screen validation, or less computation could be hidden within the A2O interval. Overall, \system{}'s latency benefit would depend on the opportunities for token reuse and early inference exposed by each workload.

\section{Related Work}
\label{sec:related_work}

Prior agent systems reuse interaction history through relatively coarse-grained forms of behavioral or planning knowledge. AppAgent~\cite{appagent} and AutoDroid~\cite{autodroid} derive app-specific operational knowledge from exploration, while AutoDroid-v2~\cite{autodroidv2} trains an SLM on app documentation and synthesized task--script pairs. MobileGPT~\cite{mobilegpt}, Agent Workflow Memory~\cite{awm}, AutoRPA~\cite{autorpa}, and Darwinian Memory~\cite{mi2026darwinian} reuse procedural units such as subtasks, workflows, RPA functions, or multi-action trajectory segments. Executable Agentic Memory~\cite{qin2026executable} stores actions in an offline-constructed graph and composes them into executable paths, whereas LLM-Explorer~\cite{llmexplorer} focuses on improving exploration itself. 
Across them, prior interactions are organized into relatively coarse-grained knowledge or procedural structures, such as app knowledge, subtasks, workflows, or executable procedures. Many of them also rely on offline exploration, model training, or memory construction. 
In contrast, \system{} reuses finer-grained execution records online, which allows \system{} to capture reuse opportunities without requiring a complete task repeat or offline preparation.

Another line of work accelerates agent pipelines or VLM inference~\cite{vdroid,vlmcache,agentx,fastv,showui}. For instance, V-Droid~\cite{vdroid} replaces autoregressive action generation with batched verification over a discrete action space, but requires a separately trained 8B verifier and still invokes a cloud LLM to update working memory. Agent-X~\cite{agentx} combines prompt restructuring, prefix caching, and N-gram speculative decoding for text-based, function-calling agents built on TinyAgent~\cite{tinyagent}. FastV~\cite{fastv} reduces visual-token processing by pruning low-importance tokens in deeper decoder layers based on early-layer attention. A recent preprint, MobileExplorer~\cite{mobileexplorer}, performs online UI exploration during VLM inference that improves task success rates and reduces latency. In summary, prior systems reuse interaction history as behavioral knowledge, optimize particular model or pipeline stages, or gather additional UI context to improve subsequent decisions. \system{} instead treats records accumulated during ordinary task executions as fine-grained computational experience, reusing it to reduce action-step latency while maintaining task success rates. It requires no additional changes to the underlying VLM backbone or agent policy.

\section{Conclusion}

This paper presents \system{}, a system that turns computational experience from prior GUI-agent executions into fine-grained inference reuse through validated token-sequence proposals, next-screen-guided early inference, and cross-step KV-state carryover. Across multiple VLMs and server, edge, and mobile platforms, \system{} reduces action-step latency while maintaining task success, without task-specific offline exploration or changes to the agent policy. This points toward on-device agents that grow more efficient as computational experience accumulates across task executions.

\bibliographystyle{ACM-Reference-Format}
\bibliography{references}


\end{document}